# BanglaNLP at BLP-2023 Task 2: Benchmarking different Transformer Models for Sentiment Analysis of Bangla Social Media Posts

**Saumajit Saha**
saha.saumajit@gmail.com

**Albert Nanda**
albert.nanda@gmail.com

## Abstract

Bangla is the 7th most widely spoken language globally, with a staggering 234 million native speakers primarily hailing from India and Bangladesh. This morphologically rich language boasts a rich literary tradition, encompassing diverse dialects and language-specific challenges. Despite its linguistic richness and history, Bangla remains categorized as a low-resource language within the natural language processing (NLP) and speech community. This paper presents our submission to Task 2 (Sentiment Analysis of Bangla Social Media Posts) of the BLP Workshop. We experimented with various Transformer-based architectures to solve this task. Our quantitative results show that transfer learning helps in better learning of the models in this low-resource language scenario. This becomes evident when we further finetuned a model that had already been finetuned on Twitter data for sentiment analysis task and that finetuned model performed the best among all other models. We also performed a detailed error analysis, finding some instances where ground truth labels need to be looked at. We obtained a micro-F1 of 67.02% on the test set and our performance in this shared task is ranked at 21 in the leaderboard.

## 1 Introduction

Sentiment analysis is the task of determining the attitude or opinion expressed in a piece of text. Typically sentiment categories are of three types: Positive, Negative, and Neutral. In today's increasingly interconnected world, where digital communication abounds, sentiment analysis has emerged as a vital component of natural language processing (NLP) and computational linguistics. It enables us to gauge public sentiment on diverse topics, monitor social media trends, and make data-driven decisions in various domains, including marketing, customer service, and politics. Social media provides an interesting platform to study sentiment analysis. People have diverse opinions regarding any topic and they express them accordingly. Mining sentiments from them often become very critical due to the trending social media lingo.

The use of slang, informal language, and emojis in social media posts can further complicate the task of sentiment analysis. The scarcity of resources and research initiatives dedicated to Bangla sentiment analysis can be attributed to several factors. Firstly, Bangla is considered a low-resource language within the NLP and speech community, primarily due to limited and scattered research efforts undertaken by individual researchers or small teams. Secondly, the development of robust deep learning models pre-trained on monolingual bengali data is not that widely available like we have numerous models pre-trained on English data.

## 2 Related Works

Early work on sentiment analysis in Bangla relied on lexicon-based and rule-based methods like in (Chowdhury and Chowdhury, 2014). Lexicon-based methods use a dictionary of sentiment words to identify the sentiment of a text. Rule-based methods use a set of rules to identify sentiment words and phrases. However, with the advancement of deep learning models, these approaches were outperformed by them because they are more capable of understanding the contextual meaning of the sentence and they do not require handcrafted rules or a set of lexicons to identify the sentiment present in text segment.

Bhowmick and Jana (2021) performed sentiment analysis using Bert and XLM-Roberta on three datasets - Prothom Alo (Islam et al., 2020), YouTube-B (Sazzed, 2020) and Book-B (Hossain et al., 2021). Kabir et al. (2023) introduced a large-scale Bangla dataset for sentiment analysis from book reviews. Islam et al. (2023) introduced a multi-domain Bangla sentiment analysis dataset across 30 different domains.

## 3 System Description

This section describes our system which is developed to classify sentiment present in Bangla social media posts. This section starts with the shared task description, followed by the description of the dataset released by the shared task organizers, then our proposed architecture which produced our team's standing on the leaderboard, and finally the results achieved and observations made. All the codes and datasets used for performing the experiments are available in https://github.com/Saumajit/BanglaNLP/tree/main/Task_2.

### 3.1 Shared Task Description

The objective of this shared task[1] (Hasan et al., 2023a) is to identify the sentiment associated with a given text segment. Given a Bangla text segment, the output produced by the system should belong to one of the 3 classes - *positive*, *negative*, and *neutral*.

### 3.2 Dataset Description

Table 1 shows a sample sentence from the given dataset for each of the 3 sentiment categories. The dataset under consideration in this shared task combines data from two distinct sources: MUBASE (Hasan et al., 2023b) and SentNob (Islam et al., 2021). The SentNob dataset consists of public comments from various social media platforms related to news and video content. These comments are curated from 13 diverse domains such as politics, education, and agriculture. On the other hand, the MUBASE dataset is a large collection of multi-platform dataset that includes manually annotated Tweets and Facebook posts, each labeled with their respective sentiment polarity. Table 2 highlights the count of positive, negative, and neutral sentences across train and development splits of the dataset respectively.

We find that almost 80% of the sentences across train and development sets have less than 20 words for each of the three sentiment categories. We illustrate this analysis in the appendix.

### 3.3 Our Approaches

We have performed several experiments by using different transformer (Vaswani et al., 2017) models as well as several traditional machine learning algorithms. We report the promising approaches here and the rest of our approaches and their results are presented in the Appendix.

#### 3.3.1 Proposed Approach : Finetuning twitter-xlm-roberta-base-sentiment[2]

Barbieri et al. (2021) pretrained xlm-roberta-base[3] model from scratch on the tweet data. The tweets were from diverse languages as they did not want to focus on any specific language. Then they finetuned their pre-trained language model on a multilingual Sentiment Analysis dataset using adapter technique (Pfeiffer et al., 2020).

We use their finetuned model checkpoint as released in Hugging Face and further finetune it on our dataset. Since this model is already well aware of multilingual linguistic features, it performs the best on this shared task compared to all the other models that we have experimented with. Pre-existing knowledge of multilingual sentiment analysis might have helped the model in better transfer learning on our data during finetuning.

We used a learning rate of $5e - 5$, AdamW (Loshchilov and Hutter, 2017) as optimizer and a batch size of 32. We used V100 GPU for finetuning. With EarlyStopping, our best finetuned model was obtained after 2 epochs and the time taken for finetuning it on our dataset was approximately 1 hour. For finetuning the transformer-based models for SequenceClassification, we had used AutoModelForSequenceClassification class from Hugging Face throughout this paper, unless otherwise specified. During the development phase of this shared task, this finetuned model gave the best performance on the *dev_test* data split. We therefore used this model for inference on the test set released by the shared task organizers.

#### 3.3.2 Other Approaches

Two other interesting models and approaches, which lie just behind our proposed approach in terms of performance, are discussed here.

1. **Finetuning BanglaBERT**[4] Sarker (2020) proposed BanglaBERT by pretraining base ELECTRA (Clark et al., 2020) model with the Replaced Token Detection objective. Their pretraining data consists of web-crawled data and post-filtering to include only bengali data from crawled webpages. We finetuned BanglaBERT on this shared task's dataset using the learning rate of $5e-5$, AdamW as an optimizer, batch size of 32, and the number of epochs as 10.

---
[1]https://github.com/blp-workshop/blp_task2
[2]https://huggingface.co/cardiffnlp/twitter-xlm-roberta-base-sentiment
[3]https://huggingface.co/xlm-roberta-base
[4]https://huggingface.co/sagorsarker/bangla-bert-base

| Sentence | Sentiment |
|---|---|
| টানা দুই হারের পর জয়ের স্বাদ পেল ইউভেন্তুস । | Positive |
| করোনায় আক্রান্ত হয়ে আরো ১ জনের মৃত্যু | Negative |
| চিন্তা করেন যারা বক্তব্যে দিচ্ছে তাদের কণ্ঠ ও ছবি দেখাতে সাহস ও পায় না | Neutral |

Table 1: Sample data for each of the three sentiment categories.

|  | Train | Dev |
|---|---|---|
| **Positive** | 12364 | 1388 |
| **Negative** | 15767 | 1753 |
| **Neutral** | 7135 | 793 |

Table 2: Dataset statistics

2. **P-Tuning XLM-Roberta-Large**[5] Models having billions of parameters often suffer from poor transferability. Yue et al. (2020) discussed that these models are too large to memorize the finetuning samples. Liu et al. (2021) introduced P-tuning, a technique which does not change the pre-trained models' parameters but evoke the stored knowledge by finding a better continuous prompt. In finetuning, all the models' parameters get updated. However in P-tuning, the parameters corresponding to continuous prompt get updated but these parameters are of several magnitude orders smaller than the pre-trained models' parameters. The advantage of P-tuning over discrete prompts is that P-tuning helps us to find better continuous prompts beyond the original vocabulary of the pre-trained language model. We used P-tuning on XLM-Roberta-Large for the sentiment classification task. We used the learning rate of $1e-4$, the number of epochs set to 15, and the batch size set to 8. This approach trained only 42.86% of the model parameters thereby saving compute and time without impacting model performance to a great extent.

---
[5]https://huggingface.co/xlm-roberta-large

### 3.4 Results and Findings

This subsection highlights the results we had obtained during the development phase of this shared task, the metric we used for evaluating model performance, results, and error analysis on the test set.

| Approach | Model | Micro-F1 |
|---|---|---|
| FT | twitter-xlm-roberta-base-sentiment | **0.68** |
| FT | BanglaBERT | 0.65 |
| PT | xlm-roberta-large | 0.63 |

Table 3: Performance of different models on the development set. FT : Finetuning, PT : P-Tuning.

#### 3.4.1 Evaluation Metric

The evaluation metric for this shared task is micro F1. Micro F1 calculates metrics globally by counting the total number of true positives, false negatives, and false positives.

#### 3.4.2 Performance on Development and Test Set

Table 3 highlights the performance of our approaches on the given dataset during the development phase. We see that the *twitter-xlm-roberta-base-sentiment* model performed the best in terms of evaluation metrics. This might have happened due to transfer learning (Farahani et al., 2021) which aims to benefit pre-trained models that need to be further trained on low-resource languages. We also finetuned *BanglaBERT*, a monolingual model, to evaluate how it performs in comparison to the other models. We see that there is a gap in its performance and that may be attributed to the monolingual nature of a model trained on a low-resource language. Finally our P-tuning approach on *xlm-roberta-large* gave a competitive performance with the above models with less number of trainable parameters. On the test set shared with us by the organizers, we obtained a micro F1 of 67.02% using our finetuned *twitter-xlm-roberta-sentiment* model. We therefore observe that the model performance slightly (∼ 1%) drops on the evaluation phase test set compared to the development phase. This helps us to understand that our finetuned model also generalized well to unseen data and thus is fairly stable in nature.

| Sentence | Ground Truth | Prediction |
|---|---|---|
| সিরিয়ায় অবস্থান করা বিদেশি বাহিনীর সমালোচনা করেছেন পুতিন । | Positive | Negative |
| আজ আইন এই রকম বলেই দিন তো দিন বেড়ে যাচ্ছে দর্শন । তাই এই বিষয় দল , বল খোঁজবেন না কঠিন শাস্তি দিবেন । | Positive | Negative |
| ভারতীয় ব্যাটসম্যানদের দাঁড়াতেই দেয়নি ইংলিশ বোলাররা । | Positive | Negative |
| মহাকাশে কি এলিয়েন আছে ? | Positive | Neutral |
| ভাগ্যরেখা অনুযায়ী আপনার আজকের দিনটি কেমন কাটতে পারে ? | Positive | Neutral |
| খুব বিরক্তিকর একটা জিনিস । খুলতে গিয়ে টাকা ছিঁড়ে যায় । | Neutral | Negative |
| আশা করেছিল ড্রন থেকে ফুলের তুরা , বুকে , পরবে ভুলে বোমা পারে গেছে | Neutral | Negative |
| বিষয়টা বেশ হাস্যরস সৃষ্টি করে | Neutral | Positive |
| ছবিটা চমৎকার ভাবে এডিট করা হইছে | Neutral | Positive |
| এবার হয়তো আপনাদের তালিবদের দুঃসহ বেদনাটা একটু কমবে আশা করি | Negative | Positive |
| সহজে বলতে গেলে শাক দিয়ে মাছ ঢাকা হচ্ছে এই আরকি | Negative | Positive |
| হিসাব টা কিভাবে বের করলেন ব্রো | Negative | Neutral |
| শুধুই জাতীয় বিশ্ববিদ্যালয় ফোকাস করছেন কেনো পাবলিক বিশ্ববিদ্যালয়ের বেকারেরর সংখ্যা তুলে ধরুন | Negative | Neutral |

Table 4: Samples where model predictions look good but ground truths look incorrect.

### 3.4.3 Error Analysis on Test set

While visually analyzing the model predictions, we find that there are several instances where our model had predicted the correct sentiment class while the corresponding ground truth labels do not seem to be correct. Table 4 shows some of the samples where our model's predictions actually look correct but ground truth annotations look incorrect. Inspite of incorrect ground truths, the model through its prior knowledge (both from transfer learning as well as finetuning on our data) was able to correctly predict the output which looks far more realistic. This stable nature of the model will help to improve data quality and get tagged data by using it to create weak sentiment labels on unseen data and then get them verified by a human-in-the-loop (Wu et al., 2021) setting.

Figure 1 denotes the confusion matrix we got by our model's predictions on the test set in the evaluation phase. We found that 67.78% positive sentences, 78% negative sentences, and 37% neutral sentences have been predicted correctly. We also found that neutral sentences got misclassified the most into positive and negative classes. Intu-itively, this could happen due to the availability of less number of neutral samples in the training data in comparison to positive and negative samples. From the dataset distribution in Table 2, we observe that the higher the number of samples seen during training, the less the number of samples getting incorrectly predicted by the model.

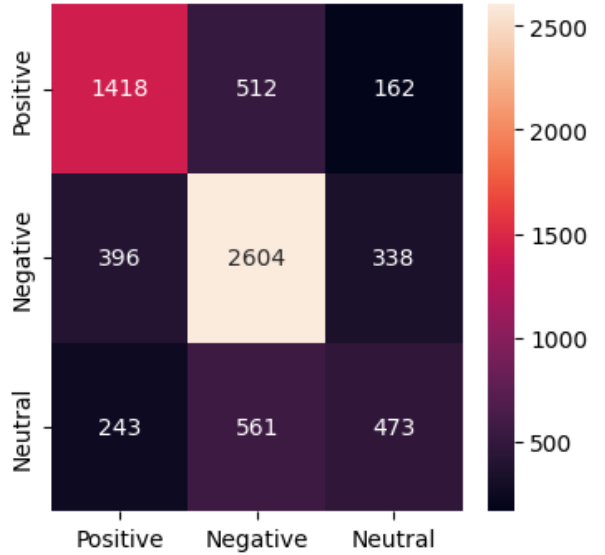

Figure 1: Confusion Matrix obtained for the test set.

## 4 Conclusion

We have provided an overview of how some of the promising approaches using transformer-based models perform with Bengali text data. We have also pointed out a few flaws in the annotation quality of the data, which if corrected, may lead to better performance of the models. We find that a transfer learning-based approach with a multilingual model works best in such a low-resource scenario when there are not too many models available that are pre-trained on a huge corpus of monolingual data. An interesting future research direction seems to be the application of recently released Large Language Models (LLMs) in the NLP space and see how they perform with a low-resource language like Bengali.

## 5 Limitations

The experiments performed, models chosen, and results that have been discussed here are purely based on a low-resource language like Bangla and the particular dataset shared for use in the Shared Task. All experiments are mostly run in v100, T4 GPU, and rarely in A100 using Google Colab. Recently released Large Language Models and ChatGPT are not used here due to compute and pricing constraints.

## 6 Appendices

In this section, we report the word count analysis per sentence across the train and development dataset. We also report some of the additional experiments we had done, which did not give satisfactory outcomes.

### 6.1 Word count distribution

Figure 2 analyzes the number of sentences which lie in the different word count intervals. For all the categories of sentiment, we find that the majority of the data samples have less than 20 words across both the train and development splits of the dataset.

### 6.2 Other experiments

Before moving to using deep learning models, we had also initially tried out several traditional machine learning algorithms like *Logistic Regression*, *Multinomial Naive Bayes* (Kibriya et al., 2005), *SGD classifier*, *Majority Voting* (Lam and Suen, 1997) of previous three classifiers and *Stacking* with XGBoost (Chen and Guestrin, 2016) as the final classifier. We had used TF-IDF (Ramos, 2003) vectorization to convert words into a vectorized representation before passing them into these classification algorithms for sentiment classification.

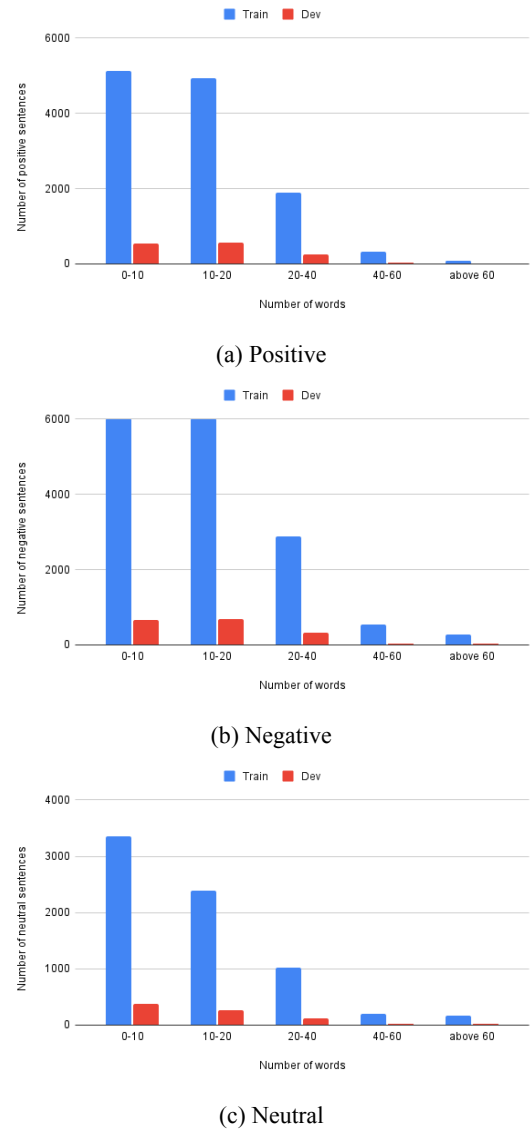

(a) Positive

(b) Negative

(c) Neutral

Figure 2: Analysis of number of sentences to the number of words present in each sentence across train and development dataset.

| Approach | Model | Micro-F1 |
|---|---|---|
| Traditional ML | Logistic Regression | 0.55 |
| | Multinomial Naive Bayes | 0.56 |
| | SGD classifier | 0.47 |
| | Majority Voting of above 3 | 0.55 |
| | Stacking | 0.54 |
| Finetuning | Bert-base-multilingual | 0.64 |
| Finetuning | Flan-t5-base | 0.47 |

Table 5: Additional Experiments

Table 5 presents the findings achieved in this task with the mentioned algorithms during the development phase.

We also used *bert-base-multilingual*[6] model to check how it performs on our task. Since it was pretrained on top of 104 languages, our intuition behind trying out this model is that the linguistic features learned by the model across different languages may help in performing our task better. From Table 5, we observe that finetuning this model gives a competitive performance.

We also tried to instruction-finetune flan-t5-base[7] model on our task. The result in Table 5 does not look promising as we have just tried to experiment with it using only a fixed setting of hyperparameters. We do not do any hyperparameter optimization here due to compute constraints. We use a learning rate of $3e-4$, batch size of 32, and number of epochs set to 5. We prepend the prompt (পাঠ্য অংশের অনুভূতি শ্রেণীবদ্ধ করুন:) to the input text to finetune the flan-t5-base model. This particular approach generates the ground-truth class label instead of classifying it into one of the pre-defined class labels which happens in a multi-class classification setting.

---

[6]https://huggingface.co/bert-base-multilingual-cased

[7]https://huggingface.co/google/flan-t5-base